\documentclass[letterpaper, 10pt, conference]{ieeeconf}      % Use this line for a4 paper

\IEEEoverridecommandlockouts                              % This command is only needed if 
\usepackage{graphics} % for pdf, bitmapped graphics files
\usepackage{epsfig} % for postscript graphics files
\usepackage{mathptmx} % assumes new font selection scheme installed
\usepackage{times} % assumes new font selection scheme installed
\usepackage{amsmath} % assumes amsmath package installed
\usepackage{amssymb}  % assumes amsmath package installed
\usepackage{color}
\usepackage{multirow}
\usepackage{subfigure}
\usepackage{booktabs}
\makeatletter
\let\NAT@parse\underfined
\makeatother
\usepackage[colorlinks,
            linkcolor=red,
            anchorcolor=blue,
            citecolor=green]{hyperref}

\definecolor{defcolor}{rgb}{0.36, 0.54, 0.66}
\definecolor{defpurple}{rgb}{1, 0, 1}

\title{\LARGE \bf
DDS3D: Dense Pseudo-Labels with Dynamic Threshold for Semi-Supervised 3D Object Detection
}

\author{Jingyu Li$^{1*}$, Zhe Liu$^{1*}$, Jinghua Hou$^{1}$, Dingkang Liang$^{2\dag}$
\thanks{This work was  supported by the Young Scientists Fund of the National Natural Science Foundation of China under Grant 62206103}
\thanks{* Equal contribution}% <-this % stops a space
\thanks{\dag Corresponding author (dkliang@hust.edu.cn)}
\thanks{$^{1}$School of Electronic Information and Communication, Huazhong University of Science and Technology}%
\thanks{$^{2}$School of Artificial Intelligence and Automation, Huazhong University of Science and Technology}%
}

\begin{document}

\maketitle
\thispagestyle{empty}
\pagestyle{empty}

%%%%%%%%%%%%%%%%%%%%%%%%%%%%%%%%%%%%%%%%%%%%%%%%%%%%%%%%%%%%%%%%%%%%%%%%%%%%%%%%
\begin{abstract}

In this paper, we present a simple yet effective semi-supervised 3D object detector named DDS3D. Our main contributions have two-fold. On the one hand, different from previous works using Non-Maximal Suppression~(NMS) or its variants for obtaining the sparse pseudo labels, we propose a dense pseudo-label generation strategy to get dense pseudo-labels, which can retain more potential supervision information for the student network. On the other hand, instead of traditional fixed thresholds, we propose a dynamic threshold manner to generate pseudo-labels, which can guarantee the quality and quantity of pseudo-labels during the whole training process. Benefiting from these two components, our DDS3D outperforms the state-of-the-art semi-supervised 3d object detection with mAP of 3.1\% on the pedestrian and 2.1\%  on the cyclist under the same configuration of 1\% samples. Extensive ablation studies on the KITTI dataset demonstrate the effectiveness of our DDS3D. The code and models will be made publicly available at \url{https://github.com/hust-jy/DDS3D}
\end{abstract}

%%%%%%%%%%%%%%%%%%%%%%%%%%%%%%%%%%%%%%%%%%%%%%%%%%%%%%%%%%%%%%%%%%%%%%%%%%%%%%%%
\section{INTRODUCTION}

Recently, LiDAR-based 3D Object Detection has received increasing attention in autonomous driving since it can help the car better understand the environment. To achieve satisfactory performance, most existing methods~\cite{yan2018second,liu2020tanet,shi2019pointrcnn} need massive labeled images for training. However, annotating 3D bounding boxes for each object is expensive and laborious.
A compromise solution is the Semi-Supervised 3D object Detection (SS3D) paradigm, i.e., using a large number of unlabeled data to boost the performance of a detector trained by a small amount of labeled data.

A pioneer for SS3D is 3DIoUMatch~\cite{wang20213dioumatch}, which addresses this task by a teacher-student paradigm, i.e., the student conduct detection training, and the teacher is in charge of annotating pseudo-labels for unlabeled images. 3DIoUMatch proposes a confidence-based label filtering strategy, while the confidence includes IoU estimation and classification score. Additionally, NMS-like post-processing is introduced to balance the quality and quantity of pseudo-labels. Although 3DIoUMatch achieves considerable performance, it needs to take tremendous effort to balance the effect of each component. Such a complex method gives rise to a question: \textit{Can SS3D be solved with a simple pipeline?}

\begin{figure}[t]
\centering
\includegraphics[width=1\linewidth]{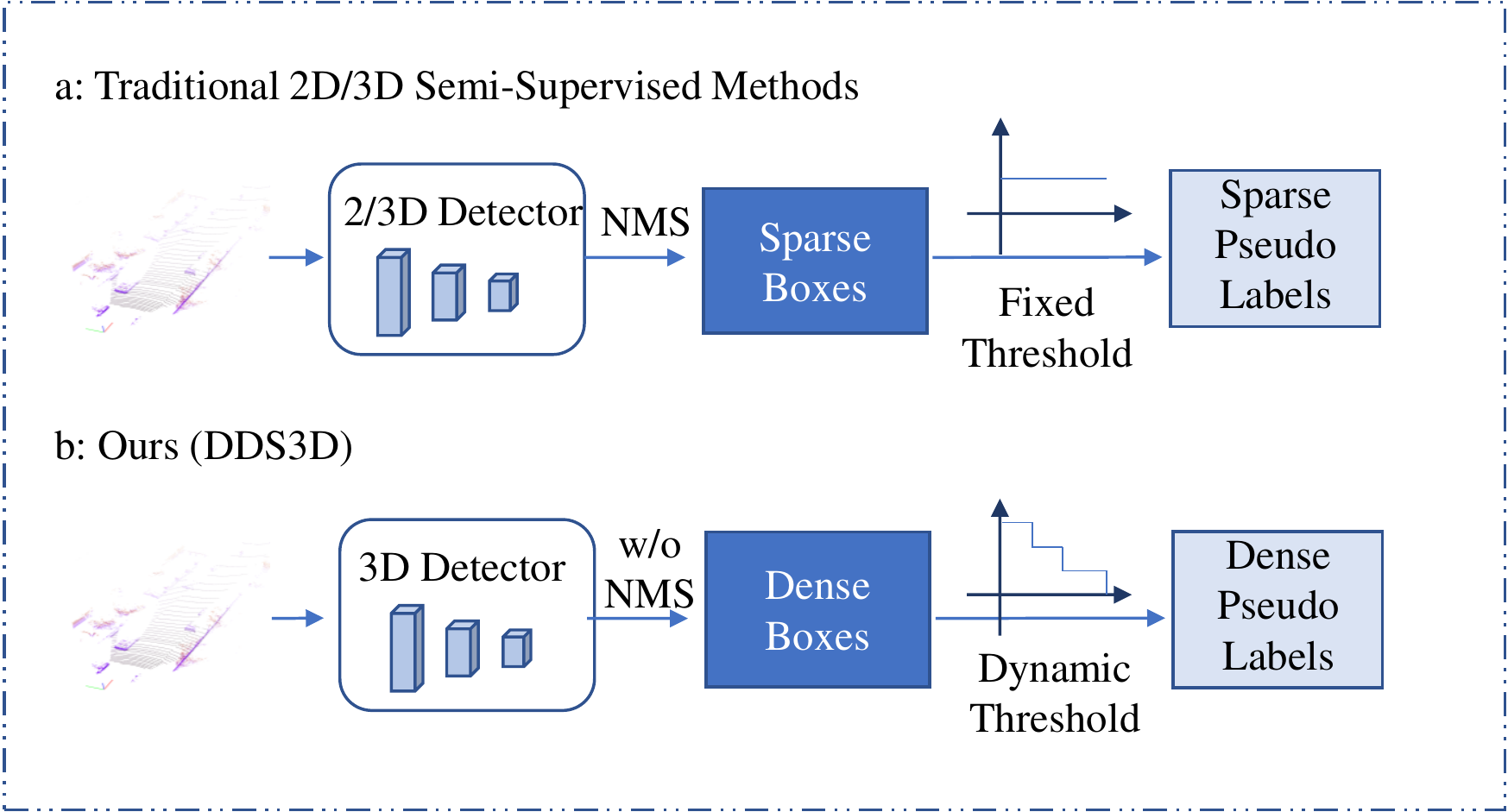}
\caption{The comparison of our method and previous methods in terms of pseudo-label generation strategy. (a) Previous methods adopt NMS to generate sparse predictions, and a fixed threshold is used to choose high-quality candidates. (b) The proposed method generates dense predictions and uses a dynamic threshold to make the model better adapted to the quality of the pseudo label.}
\label{figure:intro}
\end{figure}

To answer this question, we revisit the existing label filtering strategy and empirically find two interesting phenomena: 1) NMS is commonly used to filter out duplicate low-scoring predictions, but sparse pseudo-labels are unwise in the semi-supervised objection detection, e.g., NMS easily removes correct supervisory information; 2) Using a fixed threshold (e.g., class or IoU scores) to filter the labels is sub-optimal. Specifically, if the threshold is set to high, the massive pseudo-labels will be regarded as false negatives. Accordingly, by adopting a low fixed threshold, the model will generate massive low-quality pseudo-labels. 

In this paper, we dedicate to designing a simple yet effective semi-supervised method to circumvent the above phenomena. Specifically, 
we first remove the NMS for the teacher's predictions, which helps to generate dense pseudo-labels. As a result, the model can obtain more supervisory information. Furthermore, we propose a dynamic threshold strategy, i.e., the threshold decreases as the iteration increases.
Our intuition is that in the early stage of training, the quality of predicted results from the teacher is unstable, so using a high threshold to slowly improve the precision rate is reasonable. But in the later stages of training, the predicted results from the teacher are reliable, and using a low threshold is beneficial to boost the recall rate. We call such method Dense Pseudo-Labels with Dynamic Threshold for Semi-supervised 3D object detection (DDS3D), and Fig.~\ref{figure:intro} illustrates the difference between the proposed method and previous works in terms of generating pseudo-labels.

In summary, the main contributions of this paper are two-fold: 1) We deeply analyze the limitations of the current pseudo-label generation strategy for semi-supervised 3D object detection. Beside, we point out that the multiple components from the previous filtering label strategy are not essential; 2) We propose a simple yet effective method named DDS3D that uses dense pseudo-labels and a dynamic threshold strategy, which can well guarantee the quality and quantity of pseudo-labels during training.

Extensive experiments are conducted on the KITTI~\cite{geiger2012we} dataset, and significant improvements from the dynamic threshold strategy indicate its effectiveness. In particular, compared with state-of-the-art, when using only 1\% of labeled data, our method outperforms 3.1 absolute improvements on the pedestrian and 2.1 absolute improvements on the cyclist. 

\section{RELATED WORK}

\textbf{3D Object Detection} In the past few years, there have been numerous detection methods to deal with 3D object detection~\cite{deng2022superpoint,qi2021offboard,huang2020epnet,liu2022epnet++,liu2023stereodistill}. In terms of point cloud representations: point-based~\cite{shi2019pointrcnn,yang20203dssd,zhang2022not,zheng2021se,liu2022ss3d} and voxel-based~\cite{zhou2018voxelnet,deng2021voxel,yan2018second}. Point-based methods directly consume irregular point clouds to extract features. PointRCNN~\cite{shi2019pointrcnn} adopts PointNet++~\cite{qi2017pointnet++} as the backbone to process point clouds, binary classification to obtain foreground points and regression to generate proposals on obtained foreground points, then the proposals are refined in the second stage with the semantic and spatial information. Voxel-based methods adopt voxelization to make point clouds regular, making it possible to extract features using asymmetric functions, e.g., convolution. VoxelNet~\cite{zhou2018voxelnet} adopts 3D convolution to extract features from regular voxels. SECOND~\cite{yan2018second} replaces the conventional 3D convolution with a 3D submanifold sparse convolution to speed up. TANet~\cite{liu2020tanet} adopts a triplet attention module to make the detection model more robust by considering the feature-wise relationship. CenterPoint~\cite{yin2021center} implements the center-based method on 3D Object Detection, which considers objects as key points at the heatmap derived from the bird's-eye-view~(BEV) feature. Although the Voxel-based methods are more computationally efficient, the inevitable information loss degrades the fine-grained localization accuracy. Therefore, some methods consider how to combine point clouds and voxels. PV-RCNN~\cite{shi2020point} proposes the Voxel Set Abstraction~(VSA) that integrates multi-scale voxel features into key points, and then the keypoint features are aggregated to the RoI-grid points to learn proposal-specific features for fine-grained proposal refinement and confidence prediction.

\textbf{Semi-Supervised Learning~(SSL)} Compared to supervised learning, SSL method can only use a small amount of data, which increases the difficulty of the task. Previous works are mainly divided into two categories of methods, consistency regularization~\cite{samuli2017temporal,sajjadi2016regularization,tarvainen2017weight,rasmus2015semi,tarvainen2017mean} and pseudo-labeling~\cite{lee2013pseudo,berthelot2019mixmatch,sohn2020fixmatch}. Temporal Ensembling~\cite{samuli2017temporal} first proposes consistency regularization and many SSL methods~\cite{berthelot2019mixmatch,berthelot2019remixmatch} leverage consistency regularization. Mean Teacher~\cite{tarvainen2017mean} 
takes the teacher model as the exponential moving average~(EMA) of the student model and then adopts the consistent regularization to enforce the predictions on unlabeled data to be consistent under different data augmentations. Others adopt pseudo-labeling~\cite{lee2013pseudo},~\cite{berthelot2019mixmatch},~\cite{sohn2020fixmatch,graham20183d,zhang2021flexmatch}, which is another popular method of SSL that can be treated as a variant of consistent regularization. MixMatch~\cite{berthelot2019mixmatch} uses a series of data augmentations and applies consistency regularization on unlabeled data. FixMatch~\cite{sohn2020fixmatch} sets a confidence threshold to filter the low quality pseudo-labels. Some SSL methods~\cite{sajjadi2016regularization},~\cite{samuli2017temporal},~\cite{sohn2020fixmatch} believe that data augmentations are very important to SSL.

\textbf{SSL for 2D and 3D Object Detection} Recently, there has been lots of works~\cite{liu2020unbiased,zhou2022dense,xu2021end,sohn2020simple,li2022pseco,liu2022unbiased,tang2021humble,zhou2021instant,li2022rethinking,rizve2020defense} in 2D Semi-Supervised Object Detection~(SS-OD). Previous works have transferred a great deal of experience from SSL works to the SS-OD domain. STAC~\cite{sohn2020simple} uses Faster-RCNN~\cite{ren2015faster} as its detector and trains the teacher model with the labeled data and generates pseudo-labels on unlabeled data as a static teacher. But with the accuracy of the student model, improving static pseudo-labels might lead to the opposite effect. Unbiased Teacher~\cite{liu2020unbiased} solves the pseudo-labels bias problem caused by the class imbalance in real labels and the overfitting problem caused by the lack of labeled data. Unbiased Teacher v2~\cite{liu2022unbiased} aims at solving the ineffectiveness of the default regression supervision in semi-supervised, focusing on optimization by predicting the uncertainty of the boundary and achieving high performance under both anchor-based and anchor-free frameworks. Soft Teacher~\cite{xu2021end} proposes a soft teacher and box jitter mechanism, the former can directly assess all the box candidates from the student model, and the latter can select accurate pseudo boxes for the unlabeled regression. Dense Teacher~\cite{zhou2022dense} proposes a united form of pseudo-labels named DPL to fit the semi-supervised setting better. PseCo~\cite{li2022pseco} proposes multi-scale feature alignment, which can be regarded as a kind of data augmentation. As for 3D Semi-Supervised Object Detection~\cite{zhao2020sess,wang20213dioumatch,park2022detmatch,zhang2022atf,9812282,wei2021fgr,li2021monocular}, SESS~\cite{zhao2020sess} is the pioneer in applying the SSL framework to point-based 3D objection detection. It uses an EMA teacher and a student on top of VoteNet~\cite{qi2019deep}, asymmetric data augmentations, and three kinds of consistency losses between the predictions of the teacher and the student. 3DIoUMatch~\cite{wang20213dioumatch} proposes a multiple threshold filter strategy based on the SESS~\cite{zhao2020sess}. ATF-3D~\cite{zhang2022atf} proposes adaptive thresholds based on distance and confidence. DetMatch~\cite{park2022detmatch} jointly leverages the information of RGB images and point clouds with the Hungarian Matching algorithm~\cite{kuhn1955hungarian} to get higher performance. Compared with the above methods, our DDS3D does not require additional image information.

\begin{figure*}[htbp]
\centering
\includegraphics[width=1\linewidth]{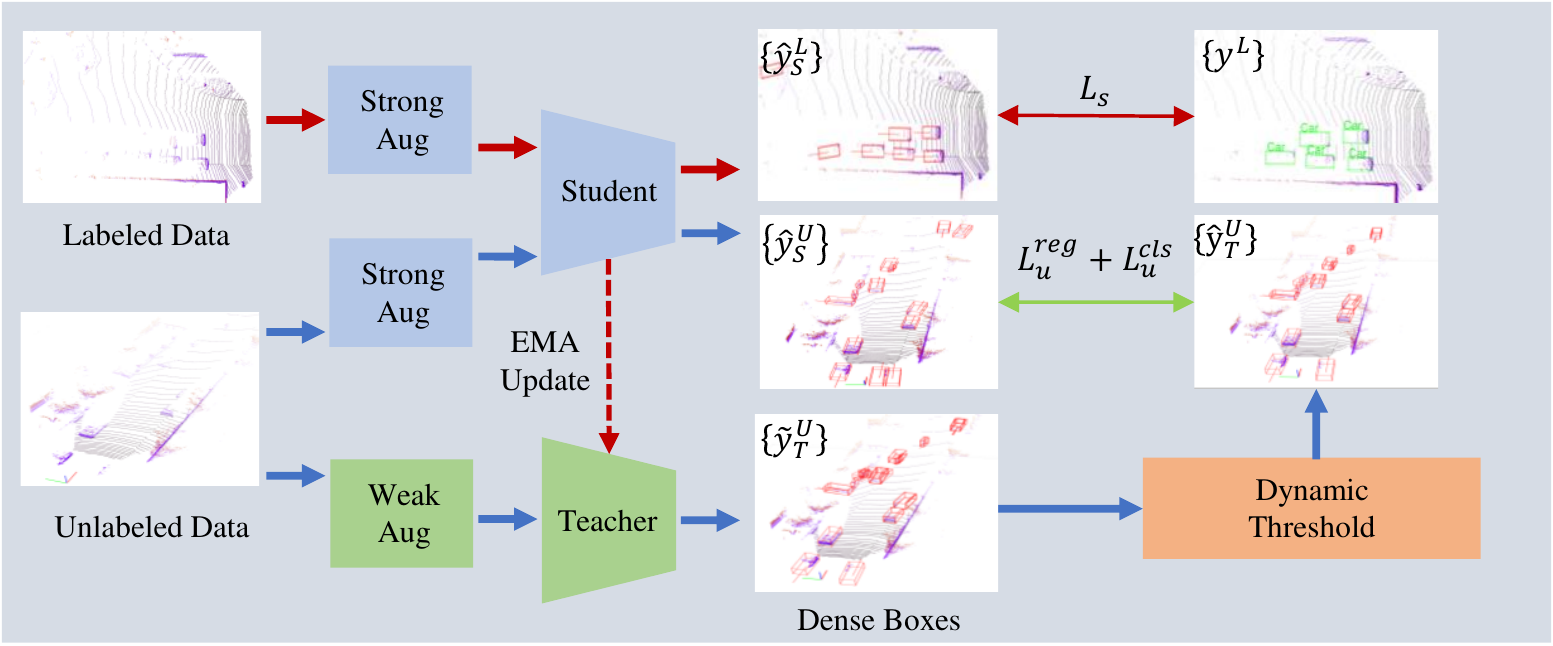}
\caption{\textbf{The pipeline of our semi-supervised framework.} For SSL, we utilize a teacher-student framework composed of a learnable student and an EMA teacher. For labeled data, the student network is supervised by GT labels directly. For unlabeled data, the student network is supervised by pseudo-labels from the EMA teacher. To get better results, we discard NMS and get dense predictions, and the gradient descent threshold is then applied to screen high-quality pseudo-labels.}
\centering
\label{fig:pipeline}
\end{figure*}
\section{METHOD}

As shown in Fig.~\ref{fig:pipeline}, we present the proposed framework of DDS3D, which contains three components: a) Teacher-Student network, b) dense pseudo-label generation, c) dynamic threshold selection. 
Before introducing the technical details of our DDS3D, we first provide the basic definitions for semi-supervised 3D object detection.
In semi-supervised 3D object detection, the total dataset includes a small part of labeled data $\left\{ {\mathop x\nolimits_i^l ,\mathop y\nolimits_i^l } \right\}_{i = 1}^{{N_l}}$ and a large amount of unlabeled data $\left\{ {\mathop x\nolimits_i^u} \right\}_{i = 1}^{{N_u}}$, where ${N_l}$ and ${N_u}$ are the number of labeled and unlabeled data, respectively. ${\mathop x\nolimits_i^l}$ and ${\mathop y\nolimits_i^l}$ represent the input point cloud data and the corresponding ground truth annotations.

\subsection{Framework of DDS3D}

\subsubsection{Teacher-student Network} 
Our DDS3D employs a teacher-student framework, where both the teacher network and the student network use the same 3D detector PV-RCNN~\cite{shi2020point} except for weight parameters and asymmetric data augmentation. More concretely, the teacher first feeds the unlabeled data with weak augmentation into the trained detector to produce the pseudo-labels, which are then utilized to supervise the student network. To ensure the effectiveness of semi-supervised learning, the pseudo-labels generated by the teacher network in the training stage are usually more accurate than the predictions of the student network. To this end, we adopt the EMA strategy during the training process.
$$
\theta _T = \alpha \theta _T + (1 - \alpha )\theta _S \eqno{(1)}
$$
where $\alpha$ is the EMA momentum and  ${\theta _T}$ and ${\theta _S}$ are the teacher and student model parameters respectively. 

\subsection{Dense Pseudo Label Generation}

\begin{figure}[h]
    \centering
    \includegraphics[width=\linewidth]{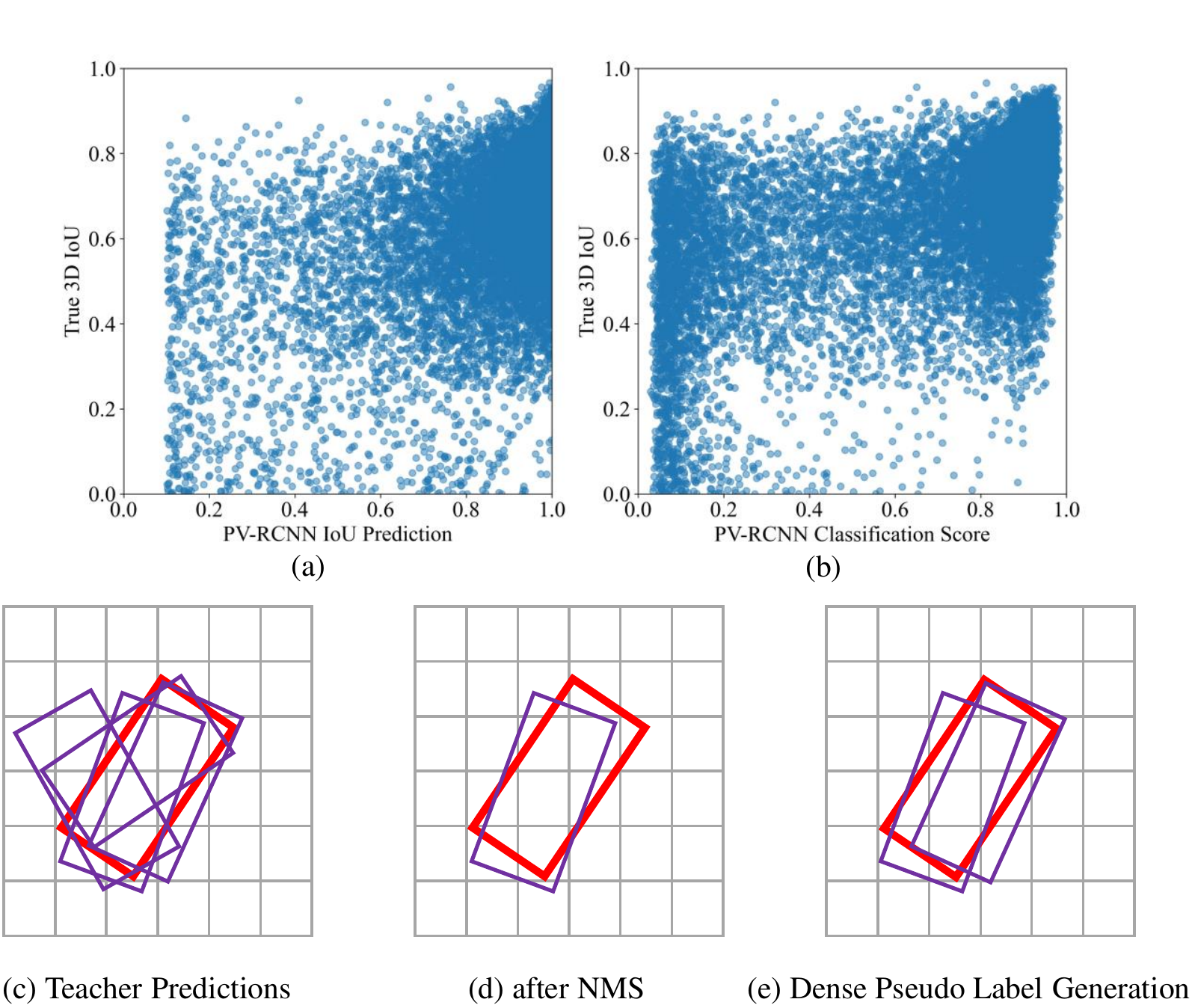}
    \caption{Comparison of classification confidence and IoU estimation with true 3D IoU on KITTI validation set, PV-RCNN is trained with 1\% labeled data; Pictorial illustration of NMS and Dense Pseudo Label Generation.}
    \vspace{-10pt}
    \label{fig:dense detials}
\end{figure}

For obtaining pseudo labels, the previous methods~\cite{wang20213dioumatch,zhao2020sess,sohn2020simple} usually introduce NMS operation to remove redundant boxes and obtain high-quality pseudo labels. However, these approaches might be sub-optimal through employing NMS to deduplicate teacher predictions since some beneficial boxes may be removed in this process. The reason for this is the inconsistency of the classification scores and the quality of regressed boxes. Fig.~\ref{fig:dense detials}~(a) and (b) illustrate the relationship among the classification score, IoU prediction score and ground-truth 3D IoU in our teacher network~(PV-RCNN). 
Although the IoU prediction is more reasonable than the classification score prediction, the prediction is still unsatisfactory compared with the ground truth. Thus, directly adopting pseudo-labels to supervise this IoU estimation branch will lead to inaccurate estimation and poor performance. 

To alleviate this problem, we propose a simple and effective dense pseudo-label generation strategy. More concretely, we retain all proposals instead of filtering out many redundant boxes by NMS operation. As a result, a ground-truth object might be detected by multiple proposals in this setting, which naturally improves the recall of detection to provide more potential supervision information than these sparse pseudo-labels obtained from NMS operation. To illustrate this process better, we provide the visualization in Fig.~\ref{fig:dense detials}, where (c) represents the proposals from the teacher.
When NMS is used to process these proposals shown in Fig.~\ref{fig:dense detials}~(d), this does not guarantee the generation of a high-quality box, which might lead to an unsatisfactory result. However, our dense pseudo label generation~(see Fig.~\ref{fig:dense detials}~(e)) can capture dense boxes to provide more potential information.

\subsection{Dynamic Threshold Selection}
\begin{figure}
    \centering
    \includegraphics[width=\linewidth]{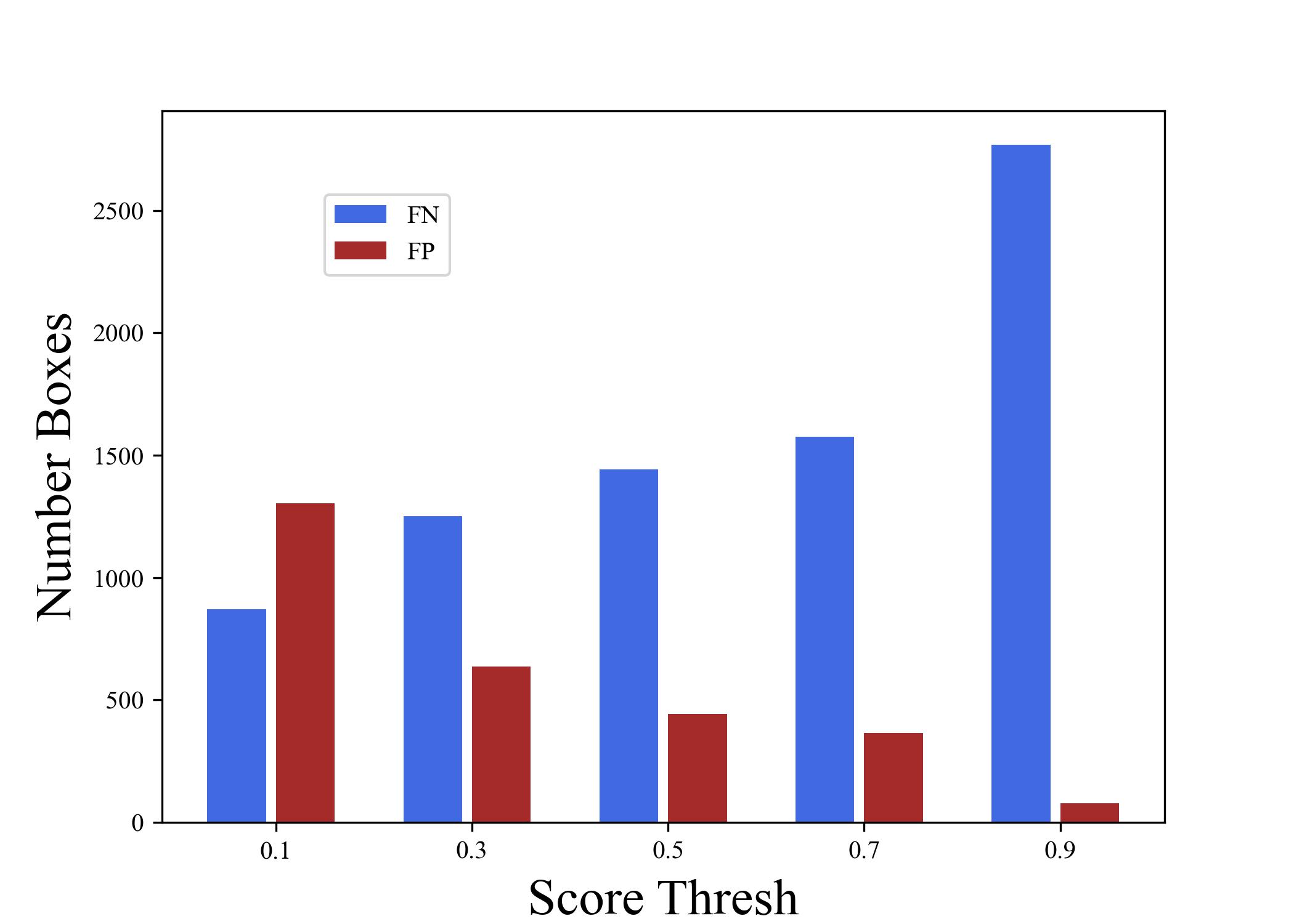}
    \caption{The number of FP and FN under different classification confidence thresholds for the KITTI validation dataset.}
    \label{fig:score}
\end{figure}

The main gap in detection performance between these two networks in the teacher-student framework is from the different EMA weights and the data augmentation strength. Although the teacher network is usually more powerful than the student network, this does not guarantee that the teacher's prediction is always more accurate than the student's. A naive method is to use a fixed threshold to filter out low-quality pseudo-labels from the teacher network. As shown in Fig.~\ref{fig:score}, the threshold is set at a high value (e.g., 0.7 and 0.9), leading to more false negative examples. Conversely, when the threshold is set at a low value (e.g., 0.1 and 0.3), the performance of the model drop drastically due to a large number of false positive. To avoid this phenomenon, different from the fixed threshold methods~\cite{wang20213dioumatch,sohn2020fixmatch}, we propose a dynamic threshold strategy. 
Specifically, we set a higher threshold to filter out most false positives to ensure the accuracy of the initial optimization direction. Then,  as the number of iterations increases, we gradually reduce the threshold to retain more potential true positives due to the stronger detection performance. Especially at the end of the training, the model usually performs better on object localization. Therefore, a lower threshold is reasonable to cover more hard objects to further boost the performance on these challenging objects. Finally, the process of the dynamic threshold selection can be formulated as:
$$
{\sigma _{cls}}(t) = \min \left( {{\sigma _{start}} - \alpha  \times \left\lfloor {\frac{t}{{steps}}} \right\rfloor ,{\sigma _{end}}} \right)
$$
where ${\sigma _{cls}}(t)$ is the threshold for classification confidence at the number of iterations $t$, ${\sigma _{start}}$ is the starting threshold, ${\sigma _{end}}$ is the end threshold, ${steps}$ is step length and $\alpha$ is the attenuation coefficient default as 0.1.

\subsection{Loss Function and Final Processing of Pseudo-Label}

In the pre-training stage, we can use a small amount of labeled data $\left\{ {\mathop x\nolimits_i^l ,\mathop y\nolimits_i^l } \right\}_{i = 1}^{{N_l}}$ to train the student network. The total loss in this stage ${L_l}$ is composed of RPN losses and RCNN losses, as:
$$
{L_l} = L_{rpn\_cls}^l + L_{rpn\_reg}^l + L_{rcnn\_iou}^l + L_{rcnn\_reg}^l \eqno{(4)}
$$
where $L_{rpn\_cls}^l$ is classification loss, $L_{rcnn\_iou}^l$ is IoU estimation loss, $L_{rpn\_reg}^l$ and $L_{rcnn\_reg}^l$ are box regression losses.

In the semi-supervised training stage, we keep the same proportion for the input labeled data and unlabeled data on each batch. For labeled data, we supervise the student with GT~(same with the pre-training stage). But for unlabeled data, given that the asymmetric data augmentation on unlabeled data for teacher and student, pseudo-labels need to go through additional geometry transformation $\mathcal{T}$ to enable the alignment with outputs of the student network.  $\mathcal{T}$ is
equal to the multiplication of the inverse weak augmentation and strong augmentation.  
Considering Iou is hard to optimize over the network, we remove the IoU estimation branch in the semi-supervised training stage. Thus, the total unsupervised loss ${L_u}$ is composed of classification loss and box regression losses, which can be computed as follows.
$$
{L_u} = L_{rpn\_cls}^u + L_{rpn\_reg}^u + L_{rcnn\_reg}^u \eqno{(5)}.
$$
Finally, the total loss $L$ for semi-supervising framework is as follows.
$$
L = {L_l} + \lambda {L_u}\eqno{(6)}
$$
Where $\lambda$ is the balance weight of the unsupervised loss.

\section{EXPERIMENTS}

\subsection{Experimental Setup}

\textbf{Dataset and Evaluation Metrics.} KITTI~\cite{geiger2012we} dataset is a common dataset for autonomous driving, which contains 7481 training samples and 7518 testing samples. Following F-PointNet~\cite{qi2018frustum}, the training samples are further divided into a \textit{train} split~(3712 samples) and a \textit{val} split~(3769 samples). For a fair comparison, We select the 1\% and 2\% labeled samples from \textit{train} split following~\cite{wang20213dioumatch} to verify the effectiveness of our method. The mean Average Precision~(mAP) with 40 recall positions is regarded as our evaluation metric. The IoU threshold for cars, pedestrians, and cyclists is set as 0.7, 0.5, and 0.5, respectively.

\textbf{Implementation Details.} We implement the basic detector PV-RCNN~\cite{shi2020point} based on the open-source framework OpenPCDet~\cite{openpcdet2020} codebase. In detail, the detection range is within [0, 70.4], [-40, 40], [-3, 1] meters along the X, Y, Z axes and the voxel size is (0.05, 0.05, 0.1) meters. The training process contains a pre-training stage and a training stage, and we train all models on 4 NVIDIA RTX 2080Ti GPUs.

For the pre-training stage, we use the labeled data to train our model for 80 epochs with batch size of 8~(default 2 samples per GPU), and we train the data ten times per epoch so that the model converges better. %we lengthen the number of traverses in each epoch to ten times the origin for the model for converging.
The detector is optimized by AdamW~\cite{loshchilov2017decoupled} optimizer with a max learning rate of 0.01.

For the training stage, we run 100 epochs with batch size of 8 for 4 GPUs~(each GPU loads 1 labeled sample and 1 unlabeled sample in each batch). Besides, we lengthen the number of traverses in each epoch to five times the origin following~\
\cite{wang20213dioumatch}. Similar to prior works~\cite{zhao2020sess,wang20213dioumatch}, we warm up the EMA momentum from 0.99 to 0.999. For our Dynamic Threshold Strategy, we set ${\sigma _{start}}$ to 0.6, ${\sigma _{end}}$ to 0.4 and ${steps}$ to 1000. Further, for a fair comparison, we adopt the same data augmentations as 3DIoUMatch, including the GT Sampling for labeled data and basic geometric transformations for unlabeled data. The basic geometric involves random flip along the \textit{X} axis, random global scaling with a scale factor sampled from $[0.95,1.05]$, and global rotation around \textit{Z} axis with a random angle sampled from $\left[ { - \frac{\pi }{4}, + \frac{\pi }{4}} \right]$.

% For the inference stage, we forward the point cloud of a scene to the student network to generate proposals. Then, we keep the top-100 proposals from 3D voxel CNN with a 3D IoU threshold of 0.7 for NMS  following the same protocol as \cite{shi2020point}. These proposals are further refined in the detection head, after refinement, NMS with IoU threshold of 0.1 is further applied to remove the redundant predictions.

\subsection{Results on KITTI}
As shown in Table \ref{all_result}, we provide a comparison with the superior semi-supervised 3D object detection method 3DIoUMatch under the settings of 1\% and 2\% labeled data on the KITTI~\cite{geiger2012we} \textit{val} split. For a fair comparison, we use the same detector PV-RCNN~\cite{shi2020point} as the labeled-data-only baseline. In Table \ref{all_result}, our method outperforms the labeled-data-only baseline by 2.7\%, 5.8\% and 8.0\% on cars, pedestrians and cyclists under 1\% labeled data. Besides, our DDS3D achieves 3.1\% and 2.1\% mAP improvement over 3DIoUMatch~\cite{wang20213dioumatch} on pedestrians and cyclists, which illustrates the superiority of our DDS3D by considering the dense pseudo-label generation and the dynamic threshold strategies. Similar conclusions for 2\% of the labeled data. For cars, our DDS3D achieves similar performance to 3DIoUMatch. The reason behind is that the detector on the category of the car has already achieved satisfactory results in the pre-training stage, which is difficult to improve by pseudo-labels in the semi-supervised framework.

% We evaluate our DDS3D . PV-RCNN*~\cite{shi2020point} is our labeled-data-only baseline. As shown in Table \ref{all_result}, our method outperforms the labeled-data-only baseline by 2.7\%, 5.8\% and 8.0\% on cars, pedestrians and cyclists under 1\% labeled data. Besides, our DDS3D achieves 3.1\% and 2.1\% mAP improvement over 3DIoUMatch~\cite{wang20213dioumatch} on pedestrians and cyclists, respectively. For 2\% labeled data, our DDS3D outperforms 3DIoUMatch* with mAP of 3.9\% on pedestrians. Our method DDS3D achieves similar performance to 3DIoUMatch on car. We believe that the reason is that the car has already achieved good results in the pre-training stage, and it's difficult to improve by pseudo-labels.

\begin{table}[t]
\small
\centering
\caption{The Performance for \textbf{Car}, Pedestrian and Cyclist on the KITTI \textbf{\textit{val}} set with different learned ratios. \textbf{Ped.} and \textbf{Cyc.} are short for Pedestrian and Cyclist. * denotes the reproduced results.}
\begin{tabular}{l|c|c|c|c|c|c}
\toprule
\multirow{2}{*}{Method} 
& \multicolumn{3}{c|}{1\%} 
& \multicolumn{3}{c}{2\%} \\
\cline{2-7}
& Car & Ped. & Cyc. & Car & Ped. & Cyc. \\
\midrule
PV-RCNN & 73.5 & 28.7 & 28.4 & 76.6 & 40.8 & 45.5 \\
PV-RCNN* & 73.3 & 29.0 & 30.5 & 76.5 & 43.2 & 43.4 \\
3DIoUMatch & \textbf{76.0} & 31.7 & 36.4 & 78.7 & 48.2 & \textbf{56.2} \\
3DIoUMatch* & \textbf{76.0} & 30.7 & 36.8 & 78.6 & 45.3 & 53.0 \\
\midrule
DDS3D~(Ours) & \textbf{76.0} & \textbf{34.8} & \textbf{38.5} & \textbf{78.9} & \textbf{49.4} &53.9 \\
\bottomrule
\end{tabular}
\label{all_result}
\end{table}

\subsection{Ablation Study}

\begin{figure*}[t!]
\centering
\includegraphics[width=1\linewidth]{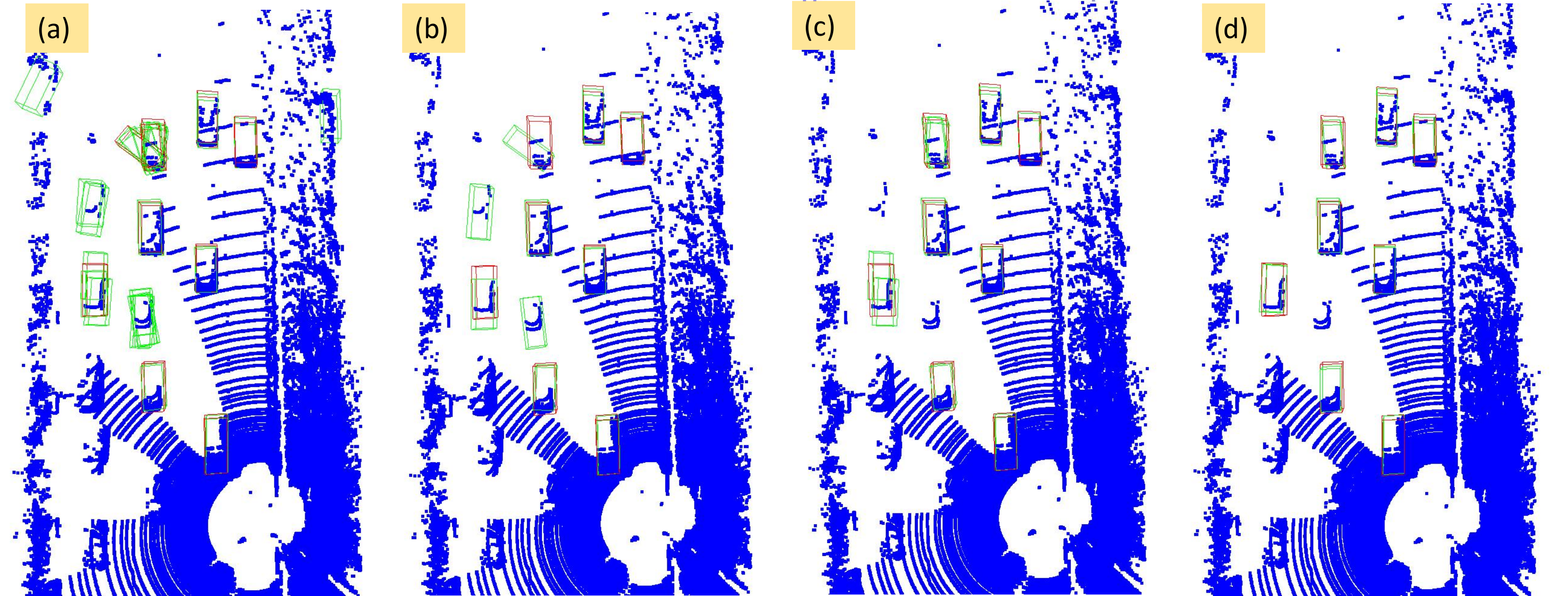}
\caption{\textbf{The Visualization of detection results on the KITTI dataset. (a), (b), (c) and (d) denote teacher's proposals, sparse pseudo-labels, dense pseudo-labels and prediction results. The prediction and GT are in green and red, respectively.}}
\centering
\label{fig:vis}
\end{figure*}

We present ablation studies with 1\% labeled data to analyze the effectiveness of our proposed components in DDS3D on KITTI~\cite{geiger2012we} \textit{val} split. Table~\ref{ablation} summarizes the ablation results on our dense pseudo-label generation mode~(DPLG) and dynamic threshold module~(DT). We adopt the labeled-data-only PV-RCNN as the baseline~(Exp (a)  of Table~\ref{ablation}). To validating the effectiveness of the dynamic threshold filter, we add a fixed threshold~(FT) manner as a comparison. SSL stands for paradigm using semi-supervised learning. 

%Compared to PV-RCNN~\cite{shi2020point}, the naive pseudo-label method (the Exp (b) of Table~\ref{ablation}) only achieves 1\% and 3\% mAP improvement on pedestrians and cyclists. Further, a fixed threshold filter brings consistent improvement (the Exp (c) of Table~\ref{ablation}), demonstrating that generating pseudo-labels by threshold filter can boost performance by filtering out low-quality prediction results.

% \begin{table}[h]
% \small
% \centering
% \caption{The ablation of the improvement of each component on the KITTI \textbf{val} set. SSL, FT, DT and DPLG denote Semi-Supervised Learning, Fixed Threshold, Dynamic Threshold and Dense Pseudo-Label Generation.}
% \begin{tabular}{c|c|c|c|c|c|c}
% \toprule
% SSL & FT & DT & DPLG & Car & Pedestrian & Cyclist \\
% \midrule
% & & & & 73.3 & 29.0 & 30.5 \\
% \checkmark & & & & 75.0 & 30.8 & 33.5 \\
% \checkmark & \checkmark & & & 75.6 & 33.3 & 34.6 \\
% \checkmark &  &\checkmark & & \textbf{76.0} & 34.2 & 34.0 \\
% \checkmark &\checkmark & &\checkmark & \textbf{76.0} & 34.4 & 36.0 \\
% \checkmark & &\checkmark & \checkmark & \textbf{76.0} & \textbf{34.8} & \textbf{38.5} \\
% \bottomrule
% \end{tabular}
% \label{ablation}
% \end{table}

\begin{table}[t]
\small
\centering
\caption{The ablation of the improvement of each component on the KITTI \textbf{\textit{val}} set. SSL, FT, DT and DPLG denote Semi-Supervised Learning, Fixed Threshold, Dynamic Threshold and Dense Pseudo-Label Generation. \textbf{Ped.} and \textbf{Cyc.} are short for Pedestrian and Cyclist.}
\begin{tabular}{c|c|c|c|c|c|c|c}
\toprule
Exp ID & SSL & FT & DT & DPLG & Car & Ped. & Cyc. \\
\midrule
a& && & & 73.3 & 29.0 & 30.5 \\
b&\checkmark & & & & 75.0 & 30.8 & 33.5 \\
c&\checkmark & \checkmark & & & 75.6 & 33.3 & 34.6 \\
d&\checkmark &  &\checkmark & & \textbf{76.0} & 34.2 & 34.0 \\
e&\checkmark &\checkmark & &\checkmark & \textbf{76.0} & 34.4 & 36.0 \\
f (ours)&\checkmark & &\checkmark & \checkmark & \textbf{76.0} & \textbf{34.8} & \textbf{38.5} \\
\bottomrule
\end{tabular}
\label{ablation}
\end{table}

% \textbf{Effect of dense pseudo label generation module} Compared with the naive pseudo-label baseline with a fixed threshold (Exp (c) of Table~\ref{ablation}), our dense pseudo-label generation (Exp (e) of Table~\ref{ablation}) achieves 0.4\%, 1.1\% and 1.4\% mAP on car, pedestrian and cyclist, respectively. There are two reasons: 1) On the one hand, some boxes with high IoU with GT but low scores are filtered out by NMS as redundant boxes; 2) On the other hand, although NMS removes many false positives, it also removes a small number of correct but low-scoring predictions. Therefore, using NMS directly to generate the pseudo labels for the teacher network might be sub-optimal. By combining DT and DPLG, i.e., Exp (f) of Table~\ref{ablation}, we achieve the best result in all category. 

\textbf{Effect of dense pseudo label generation module.} Compared with the naive pseudo-label baseline with a fixed threshold (Exp (c) of Table~\ref{ablation}), our dense pseudo-label generation (Exp (e) of Table~\ref{ablation}) achieves 0.4\%, 1.1\% and 1.4\% mAP on car, pedestrian and cyclist, respectively. The main reason is that some boxes with high IoU with GT but low scores are filtered out by NMS as redundant boxes. Our dense pseudo-label generation mode can effectively deal with this case.

\textbf{Effect of dynamic threshold.} 
For a fair comparison, we extensively search for the fixed confidence threshold, as shown in Table \ref{different thresh}. It is worth noting that different optimal thresholds might be required for different object classes. To reduce the number of hyperparameters, we use the same threshold to filter all categories. As shown in Table \ref{different thresh}, setting the fixed threshold to 0.4 is a reliable choice. Thus, we choose the value of 0.4 as the fixed threshold Table~\ref{ablation}. Compared with our baseline with the fixed threshold Table~\ref{ablation}~(exp (c)), our dynamic threshold strategy shown in Table~\ref{ablation}~(exp (d)) achieves 0.4\% mAP improvement on cars and 0.9\% mAP improvement on pedestrians. Moreover, our dynamic threshold strategy with dense pseudo label generation brings a significant improvement to the fixed threshold, which achieves 0.4\% mAP improvement on pedestrians and 2.5\% mAP improvement on cyclists. The naive pseudo-label baseline proves that directly using teachers' proposals as pseudo-labels to supervise the student model gets low-performance improvement. On the other hand, it is proven that supervision with high quality and sufficient pseudo-labels is necessary. Therefore, it demonstrates the effectiveness of our dynamic threshold and more details will be discussed in later sections.

\begin{table}[h]
\small
\centering
\caption{The ablation of the fixed and dynamic threshold.}
\begin{tabular}{c|c|c|c|c}
\toprule
Threshold & Car & Pedestrian & Cyclist & mAP \\
\midrule
0.3 & 75.0 & 31.8 & 32.8 & 46.5\\
0.4 & 75.6 & 33.3 & 34.6 & 47.8\\
0.5 & \textbf{76.2} & 29.4 & 34.9 & 46.8\\
0.6 & 74.0 & 24.7 & 30.0 & 42.9\\
DT (Ours) & 76.0 & \textbf{34.8} & \textbf{38.5} & \textbf{49.8}\\
\bottomrule
\end{tabular}
\label{different thresh}
\end{table}

\begin{table}[h]
\centering
\small
\caption{The ablation of the range of the dynamic threshold (DT).}
\begin{tabular}{c|c|c|c|c}
\toprule
Range of DT & Car & Pedestrian & Cyclist & mAP \\
\midrule
0.4$\rightarrow$0.6 & 75.7 & 31.9 & 34.5 & 47.4\\
0.6$\rightarrow$0.4 & 75.8 & 34.5 & \textbf{36.8} & 49.0\\

0.7$\rightarrow$0.3 & 75.8 & \textbf{36.2} & 35.4 & \textbf{49.1}\\
0.8$\rightarrow$0.3 & \textbf{76.2} & 35.0 & 35.7 & 49.0\\
\bottomrule
\end{tabular}
\label{different threshold strategies}
\end{table}

% \jyli{Table~\ref{different thresh strategy} shows the performance of different dynamic threshold strategies. DT represents the dynamic threshold strategy. For a fair comparison, all experiments are trained under the same 1\% settings for 60 epochs with ${step} as 500$, the results are averaged over three different sampled splits for 1\% settings. On the one hand, by comparing the first two lines, we can prove that the correct threshold change direction is from high to low. On the other hand, it can be proved that our dynamic threshold strategy is flexible and changeable by comparing the last three lines. The above experiments demonstrate our dynamic threshold is loose.}

\textbf{Effect of different threshold strategies} Table~\ref{different threshold strategies} shows the performance of our dynamic threshold strategy under different settings, and all experiments are trained for 60 epochs with ${steps}$ as 500. We design some different start and end of the dynamic threshold strategy and different threshold trends: high-to-low~(our dynamic strategy) and low-to-high. According to the result, our dynamic threshold strategy always works well no matter how we choose the start and end. Moreover, we proposed high-to-low dynamic threshold strategy brings effective improvement, but the low-to-high strategy drops the performance, which demonstrates the effectiveness of our proposed dynamic threshold strategy and our hypothesis about the SSL that in the early training stage, the teacher needs a higher threshold to filter out wrong predictions that are going to be pseudo-labels, which ensures the accuracy of the initial optimization direction, and in the latter training stage, the teacher needs lower threshold to keep more predictions, which covers more hard objects so as to further boost the performance on these challenging objects.

\subsection{Qualitative Results and Analysis}
Fig.~\ref{fig:vis} shows the visualizations of the predictions by PV-RCNN~\cite{shi2020point}, 3DIoUMatch~\cite{wang20213dioumatch}, and DDS3D with 2\% labeled data on the KITTI~\cite{geiger2012we} dataset the bird's-eye view. As shown in Fig.~\ref{fig:vis}, (a) generates the pseudo labels from the teacher network PV-RCNN before NMS operation. (b) shows the sparse predictions from 3DIoUMatch through filtering out (a) in NMS operation, whose high-quality objects are obtained by the score ranking in NMS. Thus, some high-quality but low-score boxes may be eliminated due to the inconsistency between the confidence scores and the quality of the regressed boxes. (c) represents our dense pseudo-label generation strategy, which keeps more high-quality boxes compared to sparse pseudo-labels. (d) shows the final pseudo labels produced by our DDS3D, where all objects can be accurately localized. This effectively illustrates our dense pseudo labels are more reliable than traditional sparse pseudo labels.

\section{CONCLUSION}

In this paper, we have presented a novel semi-supervised 3D object detection framework named DDS3D, which involves a dense pseudo-label generation mode and a dynamic threshold strategy. The dense pseudo label generation can retain more beneficial pseudo labels compared with the manner of filtering out a large number of redundant pseudo labels through NMS operation. Besides, the proposed dynamic threshold strategy can effectively adjust a proper threshold to cover more reliable pseudo labels, which is essential for our teacher-student semi-supervised framework.
Finally, DDS3D outperforms the state-of-the-art method on 1\% labeled data under the same settings, and the experiment results on the KITTI dataset validate the effectiveness of these two components in our DDS3D. In the future, we hope that our DDS3D can be extended to more 2D/3D semi-supervised 3D object detection frameworks.

%We leverage asymmetric data augmentations and pseudo-label filter on a teacher-student mutual learning framework to facilitate student to learn from the EMA teacher. 
% In general, we first discuss and analyze the problems about NMS and the fixed threshold. Then we remove the post-processing operation on teacher's predictions to get dense pseudo-labels, and filter pseudo-labels by a flexible dynamic confidence threshold. 

% The experiment results on KITTI dataset validate the effectiveness of our method, which achieves a significant improvement over our baseline and outperforms the state-of-the-art method on 1\% labeled data under the same settings. Moreover, our dynamic threshold can be extended to 2D detectors as the future work.

% \clearpage
{\small
\bibliographystyle{ieee_fullname}
\bibliography{reference}

\begin{thebibliography}{10}\itemsep=-1pt

\bibitem{berthelot2019remixmatch}
David Berthelot, Nicholas Carlini, Ekin~D Cubuk, Alex Kurakin, Kihyuk Sohn, Han
  Zhang, and Colin Raffel.
\newblock Remixmatch: Semi-supervised learning with distribution alignment and
  augmentation anchoring.
\newblock In {\em Proc. of International Conference on Learning
  Representations}, 2020.

\bibitem{berthelot2019mixmatch}
David Berthelot, Nicholas Carlini, Ian Goodfellow, Nicolas Papernot, Avital
  Oliver, and Colin~A Raffel.
\newblock Mixmatch: A holistic approach to semi-supervised learning.
\newblock {\em Advances in neural information processing systems}, 32, 2019.

\bibitem{deng2021voxel}
Jiajun Deng, Shaoshuai Shi, Peiwei Li, Wengang Zhou, Yanyong Zhang, and
  Houqiang Li.
\newblock Voxel r-cnn: Towards high performance voxel-based 3d object
  detection.
\newblock In {\em Proc. of the AAAI Conf. on Artificial Intelligence},
  volume~35, pages 1201--1209, 2021.

\bibitem{deng2022superpoint}
Shuang Deng, Qiulei Dong, Bo Liu, and Zhanyi Hu.
\newblock Superpoint-guided semi-supervised semantic segmentation of 3d point
  clouds.
\newblock In {\em 2022 International Conference on Robotics and Automation
  (ICRA)}, pages 9214--9220. IEEE, 2022.

\bibitem{geiger2012we}
Andreas Geiger, Philip Lenz, and Raquel Urtasun.
\newblock Are we ready for autonomous driving? the kitti vision benchmark
  suite.
\newblock In {\em Proc. of IEEE Intl. Conf. on Computer Vision and Pattern
  Recognition}, pages 3354--3361. IEEE, 2012.

\bibitem{graham20183d}
Benjamin Graham, Martin Engelcke, and Laurens Van Der~Maaten.
\newblock 3d semantic segmentation with submanifold sparse convolutional
  networks.
\newblock In {\em Proceedings of the IEEE conference on computer vision and
  pattern recognition}, pages 9224--9232, 2018.

\bibitem{huang2020epnet}
Tengteng Huang, Zhe Liu, Xiwu Chen, and Xiang Bai.
\newblock Epnet: Enhancing point features with image semantics for 3d object
  detection.
\newblock In {\em Proc. of European Conference on Computer Vision}, pages
  35--52. Springer, 2020.

\bibitem{kuhn1955hungarian}
Harold~W Kuhn.
\newblock The hungarian method for the assignment problem.
\newblock {\em Naval research logistics quarterly}, 2(1-2):83--97, 1955.

\bibitem{lee2013pseudo}
Dong-Hyun Lee et~al.
\newblock Pseudo-label: The simple and efficient semi-supervised learning
  method for deep neural networks.
\newblock In {\em Workshop on challenges in representation learning, ICML},
  volume~3, page 896, 2013.

\bibitem{li2022pseco}
Gang Li, Xiang Li, Yujie Wang, Shanshan Zhang, Yichao Wu, and Ding Liang.
\newblock Pseco: Pseudo labeling and consistency training for semi-supervised
  object detection.
\newblock In {\em Proc. of European Conference on Computer Vision}, 2022.

\bibitem{li2022rethinking}
Hengduo Li, Zuxuan Wu, Abhinav Shrivastava, and Larry~S Davis.
\newblock Rethinking pseudo labels for semi-supervised object detection.
\newblock In {\em Proc. of the AAAI Conf. on Artificial Intelligence},
  volume~36, pages 1314--1322, 2022.

\bibitem{li2021monocular}
Peixuan Li and Huaici Zhao.
\newblock Monocular 3d detection with geometric constraint embedding and
  semi-supervised training.
\newblock {\em IEEE Robotics and Automation Letters}, 6(3):5565--5572, 2021.

\bibitem{liu2022ss3d}
Chuandong Liu, Chenqiang Gao, Fangcen Liu, Jiang Liu, Deyu Meng, and Xinbo Gao.
\newblock Ss3d: Sparsely-supervised 3d object detection from point cloud.
\newblock In {\em Proc. of IEEE Intl. Conf. on Computer Vision and Pattern
  Recognition}, pages 8428--8437, 2022.

\bibitem{liu2020unbiased}
Yen-Cheng Liu, Chih-Yao Ma, Zijian He, Chia-Wen Kuo, Kan Chen, Peizhao Zhang,
  Bichen Wu, Zsolt Kira, and Peter Vajda.
\newblock Unbiased teacher for semi-supervised object detection.
\newblock In {\em Proc. of International Conference on Learning
  Representations}, 2020.

\bibitem{liu2022unbiased}
Yen-Cheng Liu, Chih-Yao Ma, and Zsolt Kira.
\newblock Unbiased teacher v2: Semi-supervised object detection for anchor-free
  and anchor-based detectors.
\newblock In {\em Proc. of IEEE Intl. Conf. on Computer Vision and Pattern
  Recognition}, pages 9819--9828, 2022.

\bibitem{liu2022epnet++}
Zhe Liu, Tengteng Huang, Bingling Li, Xiwu Chen, Xi Wang, and Xiang Bai.
\newblock Epnet++: Cascade bi-directional fusion for multi-modal 3d object
  detection.
\newblock {\em IEEE Transactions on Pattern Analysis and Machine Intelligence},
  2022.

\bibitem{liu2023stereodistill}
Zhe Liu, Xiaoqing Ye, Xiao Tan, Errui Ding, and Xiang Bai.
\newblock Stereodistill: Pick the cream from lidar for distilling stereo-based
  3d object detection.
\newblock In {\em Proc. of the AAAI Conf. on Artificial Intelligence}, 2023.

\bibitem{liu2020tanet}
Zhe Liu, Xin Zhao, Tengteng Huang, Ruolan Hu, Yu Zhou, and Xiang Bai.
\newblock Tanet: Robust 3d object detection from point clouds with triple
  attention.
\newblock In {\em Proc. of the AAAI Conf. on Artificial Intelligence},
  volume~34, pages 11677--11684, 2020.

\bibitem{loshchilov2017decoupled}
Ilya Loshchilov and Frank Hutter.
\newblock Decoupled weight decay regularization.
\newblock {\em arXiv preprint arXiv:1711.05101}, 2017.

\bibitem{park2022detmatch}
Jinhyung Park, Chenfeng Xu, Yiyang Zhou, Masayoshi Tomizuka, and Wei Zhan.
\newblock Detmatch: Two teachers are better than one for joint 2d and 3d
  semi-supervised object detection.
\newblock In {\em Proc. of European Conference on Computer Vision}, 2022.

\bibitem{qi2019deep}
Charles~R Qi, Or Litany, Kaiming He, and Leonidas~J Guibas.
\newblock Deep hough voting for 3d object detection in point clouds.
\newblock In {\em Porc. of IEEE Intl. Conf. on Computer Vision}, pages
  9277--9286, 2019.

\bibitem{qi2018frustum}
Charles~R Qi, Wei Liu, Chenxia Wu, Hao Su, and Leonidas~J Guibas.
\newblock Frustum pointnets for 3d object detection from rgb-d data.
\newblock In {\em Proceedings of the IEEE conference on computer vision and
  pattern recognition}, pages 918--927, 2018.

\bibitem{qi2017pointnet++}
Charles~Ruizhongtai Qi, Li Yi, Hao Su, and Leonidas~J Guibas.
\newblock Pointnet++: Deep hierarchical feature learning on point sets in a
  metric space.
\newblock {\em Advances in neural information processing systems}, 30, 2017.

\bibitem{qi2021offboard}
Charles~R Qi, Yin Zhou, Mahyar Najibi, Pei Sun, Khoa Vo, Boyang Deng, and
  Dragomir Anguelov.
\newblock Offboard 3d object detection from point cloud sequences.
\newblock In {\em Proc. of IEEE Intl. Conf. on Computer Vision and Pattern
  Recognition}, pages 6134--6144, 2021.

\bibitem{rasmus2015semi}
Antti Rasmus, Mathias Berglund, Mikko Honkala, Harri Valpola, and Tapani Raiko.
\newblock Semi-supervised learning with ladder networks.
\newblock {\em Advances in neural information processing systems}, 28, 2015.

\bibitem{ren2015faster}
Shaoqing Ren, Kaiming He, Ross Girshick, and Jian Sun.
\newblock Faster r-cnn: Towards real-time object detection with region proposal
  networks.
\newblock {\em Advances in neural information processing systems}, 28, 2015.

\bibitem{rizve2020defense}
Mamshad~Nayeem Rizve, Kevin Duarte, Yogesh~S Rawat, and Mubarak Shah.
\newblock In defense of pseudo-labeling: An uncertainty-aware pseudo-label
  selection framework for semi-supervised learning.
\newblock In {\em Proc. of International Conference on Learning
  Representations}, 2020.

\bibitem{sajjadi2016regularization}
Mehdi Sajjadi, Mehran Javanmardi, and Tolga Tasdizen.
\newblock Regularization with stochastic transformations and perturbations for
  deep semi-supervised learning.
\newblock {\em Advances in neural information processing systems}, 29, 2016.

\bibitem{samuli2017temporal}
Laine Samuli and Aila Timo.
\newblock Temporal ensembling for semi-supervised learning.
\newblock In {\em Proc. of International Conference on Learning
  Representations}, volume~4, page~6, 2017.

\bibitem{shi2020point}
Shaoshuai Shi, Chaoxu Guo, Li Jiang, Zhe Wang, Jianping Shi, Xiaogang Wang, and
  Hongsheng~Li Pv-rcnn.
\newblock Point-voxel feature set abstraction for 3d object detection. 2020
  ieee.
\newblock In {\em Proc. of IEEE Intl. Conf. on Computer Vision and Pattern
  Recognition}, pages 10526--10535, 2020.

\bibitem{shi2019pointrcnn}
Shaoshuai Shi, Xiaogang Wang, and Hongsheng Li.
\newblock Pointrcnn: 3d object proposal generation and detection from point
  cloud.
\newblock In {\em Proc. of IEEE Intl. Conf. on Computer Vision and Pattern
  Recognition}, pages 770--779, 2019.

\bibitem{sohn2020fixmatch}
Kihyuk Sohn, David Berthelot, Nicholas Carlini, Zizhao Zhang, Han Zhang,
  Colin~A Raffel, Ekin~Dogus Cubuk, Alexey Kurakin, and Chun-Liang Li.
\newblock Fixmatch: Simplifying semi-supervised learning with consistency and
  confidence.
\newblock {\em Advances in neural information processing systems}, 33:596--608,
  2020.

\bibitem{sohn2020simple}
Kihyuk Sohn, Zizhao Zhang, Chun-Liang Li, Han Zhang, Chen-Yu Lee, and Tomas
  Pfister.
\newblock A simple semi-supervised learning framework for object detection.
\newblock {\em arXiv preprint arXiv:2005.04757}, 2020.

\bibitem{tang2021humble}
Yihe Tang, Weifeng Chen, Yijun Luo, and Yuting Zhang.
\newblock Humble teachers teach better students for semi-supervised object
  detection.
\newblock In {\em Proceedings of the IEEE/CVF Conference on Computer Vision and
  Pattern Recognition}, pages 3132--3141, 2021.

\bibitem{tarvainen2017mean}
Antti Tarvainen and Harri Valpola.
\newblock Mean teachers are better role models: Weight-averaged consistency
  targets improve semi-supervised deep learning results.
\newblock {\em Advances in neural information processing systems}, 30, 2017.

\bibitem{tarvainen2017weight}
A Tarvainen and H Valpola.
\newblock Weight-averaged consistency targets improve semi-supervised deep
  learning results. corr abs/1703.01780.
\newblock {\em arXiv preprint arXiv:1703.01780}, 1(5), 2017.

\bibitem{openpcdet2020}
OpenPCDet~Development Team.
\newblock Openpcdet: An open-source toolbox for 3d object detection from point
  clouds.
\newblock \url{https://github.com/open-mmlab/OpenPCDet}, 2020.

\bibitem{wang20213dioumatch}
He Wang, Yezhen Cong, Or Litany, Yue Gao, and Leonidas~J Guibas.
\newblock 3dioumatch: Leveraging iou prediction for semi-supervised 3d object
  detection.
\newblock In {\em Proc. of IEEE Intl. Conf. on Computer Vision and Pattern
  Recognition}, pages 14615--14624, 2021.

\bibitem{wei2021fgr}
Yi Wei, Shang Su, Jiwen Lu, and Jie Zhou.
\newblock Fgr: Frustum-aware geometric reasoning for weakly supervised 3d
  vehicle detection.
\newblock In {\em 2021 IEEE International Conference on Robotics and Automation
  (ICRA)}, pages 4348--4354. IEEE, 2021.

\bibitem{xu2021end}
Mengde Xu, Zheng Zhang, Han Hu, Jianfeng Wang, Lijuan Wang, Fangyun Wei, Xiang
  Bai, and Zicheng Liu.
\newblock End-to-end semi-supervised object detection with soft teacher.
\newblock In {\em Porc. of IEEE Intl. Conf. on Computer Vision}, pages
  3060--3069, 2021.

\bibitem{yan2018second}
Yan Yan, Yuxing Mao, and Bo Li.
\newblock Second: Sparsely embedded convolutional detection.
\newblock {\em Sensors}, 18(10):3337, 2018.

\bibitem{yang20203dssd}
Zetong Yang, Yanan Sun, Shu Liu, and Jiaya Jia.
\newblock 3dssd: Point-based 3d single stage object detector.
\newblock In {\em Proc. of IEEE Intl. Conf. on Computer Vision and Pattern
  Recognition}, pages 11040--11048, 2020.

\bibitem{9812282}
Mao Ye, Chenxi Liu, Maoqing Yao, Weiyue Wang, Zhaoqi Leng, Charles~R. Qi, and
  Dragomir Anguelov.
\newblock Multi-class 3d object detection with single-class supervision.
\newblock In {\em 2022 International Conference on Robotics and Automation
  (ICRA)}, pages 5123--5130, 2022.

\bibitem{yin2021center}
Tianwei Yin, Xingyi Zhou, and Philipp Krahenbuhl.
\newblock Center-based 3d object detection and tracking.
\newblock In {\em Proc. of IEEE Intl. Conf. on Computer Vision and Pattern
  Recognition}, pages 11784--11793, 2021.

\bibitem{zhang2021flexmatch}
Bowen Zhang, Yidong Wang, Wenxin Hou, Hao Wu, Jindong Wang, Manabu Okumura, and
  Takahiro Shinozaki.
\newblock Flexmatch: Boosting semi-supervised learning with curriculum pseudo
  labeling.
\newblock {\em Advances in Neural Information Processing Systems},
  34:18408--18419, 2021.

\bibitem{zhang2022not}
Yifan Zhang, Qingyong Hu, Guoquan Xu, Yanxin Ma, Jianwei Wan, and Yulan Guo.
\newblock Not all points are equal: Learning highly efficient point-based
  detectors for 3d lidar point clouds.
\newblock In {\em Proc. of IEEE Intl. Conf. on Computer Vision and Pattern
  Recognition}, pages 18953--18962, 2022.

\bibitem{zhang2022atf}
Zehan Zhang, Yang Ji, Wei Cui, Yulong Wang, Hao Li, Xian Zhao, Duo Li, Sanli
  Tang, Ming Yang, Wenming Tan, et~al.
\newblock Atf-3d: Semi-supervised 3d object detection with adaptive thresholds
  filtering based on confidence and distance.
\newblock {\em IEEE Robotics and Automation Letters}, 7(4):10573--10580, 2022.

\bibitem{zhao2020sess}
Na Zhao, Tat-Seng Chua, and Gim~Hee Lee.
\newblock Sess: Self-ensembling semi-supervised 3d object detection.
\newblock In {\em Proc. of IEEE Intl. Conf. on Computer Vision and Pattern
  Recognition}, pages 11079--11087, 2020.

\bibitem{zheng2021se}
Wu Zheng, Weiliang Tang, Li Jiang, and Chi-Wing Fu.
\newblock Se-ssd: Self-ensembling single-stage object detector from point
  cloud.
\newblock In {\em Proc. of IEEE Intl. Conf. on Computer Vision and Pattern
  Recognition}, pages 14494--14503, 2021.

\bibitem{zhou2022dense}
Hongyu Zhou, Zheng Ge, Songtao Liu, Weixin Mao, Zeming Li, Haiyan Yu, and Jian
  Sun.
\newblock Dense teacher: Dense pseudo-labels for semi-supervised object
  detection.
\newblock In {\em Proc. of European Conference on Computer Vision}, 2022.

\bibitem{zhou2021instant}
Qiang Zhou, Chaohui Yu, Zhibin Wang, Qi Qian, and Hao Li.
\newblock Instant-teaching: An end-to-end semi-supervised object detection
  framework.
\newblock In {\em Proc. of IEEE Intl. Conf. on Computer Vision and Pattern
  Recognition}, pages 4081--4090, 2021.

\bibitem{zhou2018voxelnet}
Yin Zhou and Oncel Tuzel.
\newblock Voxelnet: End-to-end learning for point cloud based 3d object
  detection.
\newblock In {\em Proc. of IEEE Intl. Conf. on Computer Vision and Pattern
  Recognition}, pages 4490--4499, 2018.

\end{thebibliography}
%\bibliography{reference}
}

\end{document}